%% file: main.tex
\pdfoutput=1

\documentclass[11pt]{article}

\usepackage[]{EMNLP2023}

\usepackage{times}
\usepackage{latexsym}

\usepackage[T1]{fontenc}

\usepackage[utf8]{inputenc}

\usepackage{microtype}

\usepackage{inconsolata}
\usepackage{microtype}
\usepackage{graphicx}
\usepackage{float}
\usepackage{amsmath}
\usepackage{amsfonts}
\usepackage{subfigure}
\usepackage{algorithmic}
\usepackage{multirow}
\usepackage[ruled,vlined]{algorithm2e}
\usepackage[normalem]{ulem}

\usepackage{booktabs}
\usepackage{placeins}
\usepackage{makecell}
\usepackage{bm}
\usepackage{multirow}
\usepackage{multicol}
\usepackage{arydshln}
\usepackage{url}
\usepackage{tablefootnote}
\usepackage{array}
\usepackage{marvosym}

\title{Zero-shot Cross-lingual Transfer without Parallel Corpus}

 \author{Yuyang Zhang$^*$, Xiaofeng Han$^*$, Baojun Wang$^*$ \\
         \{zhangyuyang4, hanxiaofeng5, puking.w\}@huawei.com}

\begin{document}
\maketitle
\def\thefootnote{*}\footnotetext{These authors contributed equally to this work}\def\thefootnote{\arabic{footnote}}
\begin{abstract}
  Recently, although pre-trained language models have achieved great success on multilingual NLP (Natural Language Processing) tasks, the lack of training data on many tasks in low-resource languages still limits their performance. 
  One effective way of solving that problem is to transfer knowledge from rich-resource languages to low-resource languages. 
  However, many previous works on cross-lingual transfer rely heavily on the parallel corpus or translation models, which are often difficult to obtain. 
  We propose a novel approach to conduct zero-shot cross-lingual transfer with a pre-trained model.
  It consists of a Bilingual Task Fitting module that applies task-related bilingual information alignment; a self-training module generates pseudo soft and hard labels for unlabeled data and utilizes them to conduct self-training. We got the new SOTA on different tasks without any dependencies on the parallel corpus or translation models.
\end{abstract}

\input{content/Introduction}

\input{content/RelatedWorks}

\input{content/Approach}
\input{content/Experiments}
\input{content/Discussion}
\input{content/Conclusion}

\clearpage
\bibliographystyle{acl_natbib}
\bibliography{anthology,emnlp2020}

\end{document}

%% file: content/Introduction.tex
\section{Introduction}
Zero-shot cross-lingual transfer is an important research topic in natural language processing. It aims to transfer knowledge learned from high-resource language data to the low-resource target language. Therefore, the performance of low-resource language whose training data is difficult to obtain can be improved without any supervision signals. 

The pre-trained language models have achieved great successes in various NLP tasks \cite{Qiu2020Pretrained}. Among those models, the multi-lingual ones show strong capabilities in cross-lingual transfer and have become the basis of many mainstream methods. The cross-lingual transfer is often conducted with Pre-trained language models in three steps:
(i) Pre-training a multi-lingual model on a large-scale multi-lingual dataset.
(ii) Fine-tuning the pre-trained model on a specific task in the source language.
(iii) Using the fine-tuned model to infer the test dataset in the target language \cite{wu2019beto}.
This kind of approach performs well but still has significant limitations. The distribution of the data used in the pre-training process and fine-tuning on the downstream tasks may differ significantly, and models may not be well trained on languages with no sufficient data during pre-training. As a result, the cross-lingual transfer capabilities of the models are limited.

Since the cross-lingual transfer in pre-training may cost many resources and time, many researchers turn to enhance the capability of zero-shot cross-lingual transfer on the fine-tuning model process. For example, they introduce Code-switch \cite{qin2020cosda}, multilingual adapters \cite{zeman2008cross}, different model constructions \cite{fang2020filter} and other methods during fine-tuning. Among them, Cross-lingual Self-training is the most studied method. In this method, the model trained on the source language can infer the unlabeled corpus on the target language to get pseudo labels. And then, the model is re-trained with the pseudo-labeled data. However, most existed methods rely heavily on the translation or parallel corpus to annotate the data on the target language. It is challenging to obtain high-quality parallel corpus or machine translation models for low-resource languages, which dramatically limits the generality of these methods.

We propose a novel zero-shot cross-lingual transfer method, constructed by Bilingual Task Fitting and Self-training methods, without the need for translation models or parallel corpus. Firstly, we collect task-related corpus in the target language and mix it with the original training corpus in the source language. Then we continue training a multi-lingual pre-trained model to fit it on the mixed corpus. Secondly, we propose a novel cross-lingual self-training method to transfer knowledge from the source language to the target language. We conduct our experiments on two datasets of different types and achieve good performance. Precisely, we make contributions as follows:

\begin{itemize}

\item[1)] We introduce a novel two-stage zero-shot cross-lingual approach to transfer knowledge from the source language to the target language without using any parallel corpus or translation model.

\item[2)] We propose an iterative training method combining the usage of soft labels and hard labels, which enhances the pre-trained model's capability of cross-lingual transfer.

\item[3)] Our framework gets new SOTA on different tasks. 
\end{itemize}

%% file: content/RelatedWorks.tex
\section{Related Works}

Currently, there are two most popular research directions about cross-lingual transfer: one is to pre-train multi-lingual language models with strong capacities of cross-lingual transfer. Another is to improve the performance of cross-lingual transfer during fine-tuning pre-trained models on downstream tasks. 

\subsection{Pre-train multi-lingual language models}

Some works align the word embeddings of a new model in the target language with a pre-trained model to accomplish cross-lingual transfer \cite{Schuster2019Crosslingual, Lu2015Deep, Jawanpuria2020Geometry, Yao2018Dynamic}. 
\cite{Artetxe2018LASER} uses a single BiLSTM encoder with a shared byte-pair encoding vocabulary for all languages. It enables the model to learn a classifier with English annotated data and transfer it to other languages without modification. MBERT \cite{devlin2018bert} is a transformer-based multi-lingual language model trained on raw data of Wikipedia in 104 languages. Similar to BERT, MBERT uses Mask Language Modeling (MLM) as a training task. XLM \cite{lample2019cross} proposes a new TLM training method, which introduces parallel corpus into the pre-training process. Experiments show that XLM achieves considerable improvements on classification and translation tasks. XLM-R \cite{conneau2019unsupervised} indicates that more training data and more extensive vocabulary can help improve the transferability of multi-lingual pre-trained models. It used 2.5TB CommonCrawl data as training data and a vocabulary of 250K tokens. Info-XLM \cite{chi2020infoxlm} proposes a new training task XLCO based on the idea of Contrastive Learning \cite{chen2020asimple}. XLCO significantly improved the performance of sentence retrieval tasks. Considering the generalizability, ease of use of the model, and a fair comparison with related works, MBERT is selected as a basis in our approach.

\subsection{Cross-lingual transfer during fine-tuning}

Various methods try to improve cross-lingual performance on downstream tasks. Some works \cite{qin2020cosda, Lee2020Codeswitch} use the Code-switch method to align the embedding spaces of the source and target languages. Randomly selected words are translated into their synonyms in the target language. \cite{Zhanf2020Margin} has improved Margin Disparity Discrepancy (MDD) by deploying Virtual Adversarial Training (VAT) to minimize the margin loss of the source domain so that the domain can be more general. \cite{Li2020Unsupervised} demonstrates the enhancement of the cross-lingual transfer by Multi-lingual Warm-Start (MTL-Ws) adaptation with a knowledge distillation framework. 

Some other methods fine-tune the pre-trained models with pseudo-labels, including soft labels and hard labels. For example, \cite{wu2020single} trains the student model with soft labels generated by multiple teachers trained in different languages. It has better performance than those methods which only use one teacher. \cite{WuQ2020Enhanced} also utilizes pseudo-meta-learning to conduct the cross-lingual transfer. \cite{fang2020filter} deploys a method, which uses KL-divergence self-teaching loss and pseudo soft labels produced from translated target language texts to solve the challenge of the inefficacy of the shared labels between source and target language.
Concerning hard labels, \cite{li2021cross} proposes a method to align the representations of the same-named entities in different languages by using the parallel corpus. \cite{wu2020unitrans} proposes a method that combines self-training and translation. It uses ensemble learning to generate multi pseudo labels and gets SOTA on the CoNLL dataset. The above two methods both rely much on either translation models or parallel corpus. \cite{2021Self} discusses the impacts of manually set thresholds on selecting samples with hard labels during a self-training process. MultiFit \cite{eisenschlos2019multifit} fine-tunes the multi-lingual pre-trained language model in the source language and then infers the model with provided unlabeled corpus in the target language to get pseudo labels. After that, they train a new monolingual language model on wiki data in the target language. At last, they fine-tune the monolingual language model on the data with pseudo labels. It performs well on several datasets. However, it can only obtain pseudo labels with the help of extra parallel-corpus-trained models and costs too much to pre-train new monolingual models for every target language. Besides, collecting corpus in task domains can be very hard for some low-resource languages.

%% file: content/Approach.tex
\section{Methods}

\begin{figure*}
	\centering
	\includegraphics[width=0.90\textwidth]
	{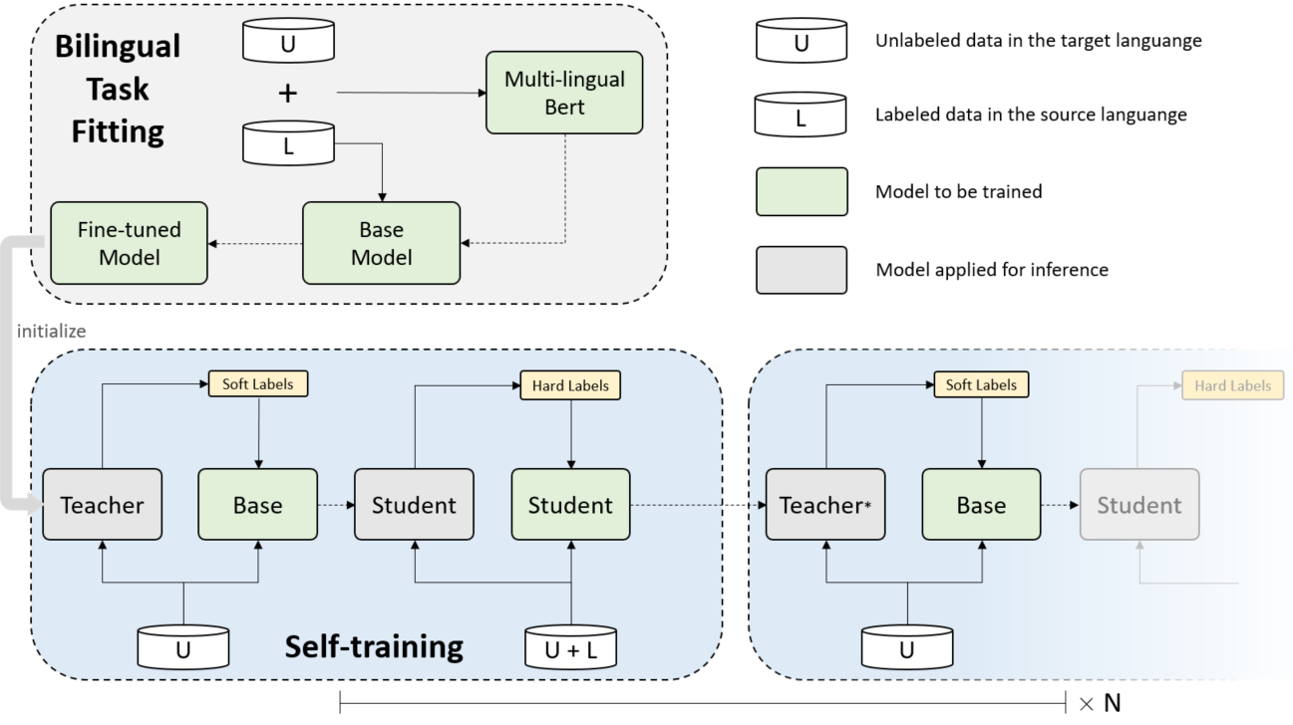}
	\caption{The framework of our proposed zero-shot cross-lingual transfer method. The Hard Labels contain labels for correctly predicted samples in corresponding source corpus and pseudo labels for target corpus. }
	\label{Fig.main_framework_v3}
\end{figure*}

We propose a novel method to conduct the zero-shot cross-lingual transfer. Figure \ref{Fig.main_framework_v3} shows our framework, consisting of two steps: Bilingual Task Fitting and Self-training. The details are shown in the following parts of this section.

\subsection{Bilingual Task Fitting}
Aiming at solving zero-shot cross-lingual problems, we assume that using only source language corpus is feasible. We use the multilingual BERT model \cite{devlin2018bert} as the foundation. This model is well-trained in more than 100 different languages. We try to improve its performance on target language with cross-lingual transfer by 1) continuing pre-training the model on MLM task with only task-oriented unlabeled corpus in the source language and target languages. We note this continue trained bilingual language model as Base Model. Furthermore, 2) fine-tuning the model with labeled task data in the source language. Since the corpus size in the source language is generally more massive than those of other languages, the mixed unlabeled training data for the MLM task are resampled. Therefore the numbers of training samples between two languages are guaranteed balanced.

For the Bilingual Task Fitting, it is critical to ensuring the unlabeled data in both source and target languages are task-oriented. For the case where task-oriented target language unlabeled corpus is unavailable, we propose a method to construct a dataset based on one assumption: the target language data are highly relevant to the task if they contain entities included in the source language training data. Therefore, we identify the entities contained in the source language training set and find their Wikipedia pages. Then we switch the language of the entities' page into corresponding ones in the target language. And finally, sentences that include these entities are extracted from Wiki corpus in the target language to conduct BTF. 

\subsection{Self-training}
In cross-lingual transfer, pseudo labels are predicted on the unlabeled target language data by the model. They are split into two types: soft labels and hard labels. Soft labels are the logits or distributions of the predicted results, while hard labels refer to manually annotated labels and the predicted classes.
Recent works \cite{tran2020cross, conneau2020unsupervised, chi2019can, eisenschlos2019multifit, Li2020PseudoLabels} have shown that utilizing the pseudo labels of the unlabeled data produces more superior results. Our framework eliminates the dependencies on any other pre-trained monolingual model or parallel corpus compared to those methods. Furthermore, high-quality pseudo labels are generated in the proposed voter's module (VM).

To be specific, we run our self-training method for N rounds. It mainly contains two steps in each round: 1) uses a Teacher Model to produce soft labels on unlabeled data in the target language, and fine-tunes the Student Model with those soft labels and Mean Squared Error (MSE) loss. In equation \ref{Eq.train_with_soft_labels}, x is the input feature of unlabeled data in the target language,  $\theta _{t}$ are the parameters of Teacher Model, and $\theta _{s}$ are the parameters of Student Model. p(x,$\theta _{t}$) and p(x,$\theta _{s}$) denotes the predicted probabilities corresponding to the input x by Teacher Model and Student Model, respectively; 2) autonomously chooses a suitable confidence threshold $\alpha$, which will be explained in the following part of this section, to select credible data in the source language and generate pseudo labels of unlabeled data in the target language. Then apply a Cross Entropy (CE) loss to fine-tune the Student Model again with the collected data. In equation \ref{Eq.train_with_hard_labels}, x is the input feature of mixed data, and y is the hard labels. The parameters in the Student Model are restored from the Base Model for each round. The parameters in the Teacher Model are restored from the Model fine-tuned with source language data in the first round and from the Student Model training with hard labels in the following rounds.

\begin{align}
    \label{Eq.train_with_soft_labels} L_{soft} & = MSE(p(x,\theta _{t}), p(x, \theta _{s})) \\
    \label{Eq.train_with_hard_labels} L_{hard} & = CrossEntropy(y, p(x, \theta _{s}))
\end{align}

In most self-training methods, the confidence threshold for filtering pseudo labels is empirically set and kept unchanged. However, distinctively, we select $\alpha$ automatically in each round by using labeled data in the source language and the models trained by soft labels. Like the grid search method, we set several thresholds and infer the model with the labeled training data in the source language. The accuracy and recall rates are computed only using samples that meet this threshold, and we choose $\alpha$ by looking for the point of the largest rate of accuracy changes, or we can say it the point where the curve has the greatest curvature. In equation \ref{Eq.threshold_choose}, $P_{recall}$ is the predicted labels meet the threshold. L means the annotated labels of the data on the source language. $t_i$ is the threshold to be selected. For the samples in training data in the source language, they will be discarded during self-training if they are not predicted correctly by the model trained by soft-labels.
\begin{equation}
\begin{aligned}
    Acc & = \frac{Num(P_{recall}\cap L)}{Num(P_{recall})} \\
    \alpha & = max(\lim_{t \to t_{i}} \frac{\nabla^2Acc} {\nabla t^2})\ \ t_{i} \in (0,1)
    \label{Eq.threshold_choose}
\end{aligned}
\end{equation}

Moreover, we introduce a new Voters Module to replace the downstream networks during self-training. The main idea behind this module is to conduct ensemble learning. Using several voters with similar architectures, we make our model more robust to the noisy data while not introducing too many parameters. 
Specifically, there are M voters in this module. Each voter consists of an independent multi-layer network as the feature encoder. There are minor differences among the dimensions of intermediate layers of the voters, though the input and output dimensions are the same. We believe that these voters can generate similar yet not identical results with the same inputs. At last, there is a decision policy on the top of VM. It considers all the outputs of voters and predicts the final pseudo labels by manual rules, as shown in Equation \ref{Eq.decision_policy}.

\begin{equation}
\begin{aligned}
    & y_{i} = softmax(W_{i}h_{i} + b_{i}) \\
    & prediction=j \quad if \quad \{y_{i}^{j}>\alpha\ and\\
    & argmax(y_{i}) = j,\quad i\in{1..M} \}
    \label{Eq.decision_policy}
\end{aligned}
\end{equation}

Here, $h_{i}$ is the output of the feature encoder $i$ and $y_{i}$ are the output logits. $W_i$ and $b_i$ are the trainable parameters. $y_{i}^{j}$ is the confidence of the sample belonging to class $j$, predicted by feature encoder $i$. $\alpha$ is the confidence threshold. 

In our policy, all the voters must take precisely the exact predictions, and all the confidences must meet the threshold $\alpha$. Then, the predicted labels are annotated as pseudo labels. Otherwise, the samples are abandoned for the next round of self-training.


%% file: content/Experiments.tex
\section{Experiments}
In the following experiments, we consider the problem as a zero-shot learning task. Therefore, annotated training data is only supported in the source language during all the training stages. In all the experiments, we employ the cased multi-lingual BERT as our baseline. 

\subsection{Dataset}
\textbf{CLS} The Cross-Lingual Sentiment (CLS) dataset is used to verify the performance of our proposed method. It is a sentiment classification dataset in 4 different languages. The English data are used as the source language, and the target languages include German, French, and Japanese. The corpus is collected from reviews of certain kinds of products from Amazon, including books, DVDs, and music. A review with more than three stars is labeled as positive and is labeled negative if less than three stars. In each language, all the corpus are split into three subsets: training, test, and unlabeled, whose sizes are 2000, 2000, and 9000-50000, respectively. We use the unlabeled data directly during the BTF and Self-training stages, similar to other works done to omit the process of collecting data.

\begin{table}[htbp]
\small
\centering
\hspace*{\fill} \\
\begin{tabular}{lccc}
\toprule
\textbf{Language} & Domain & Train/Test & Unlabeled \\
\midrule
\multirow{3}*{English (EN)}
~ & Books & 2,000 & - \\
~ & Dvd & 2,000 & - \\
~ & Music & 2,000 & - \\
\midrule
\multirow{3}*{Germany (DE)}
~ & Books & 2,000 & 165,457 \\
~ & Dvd & 2,000 & 91,506 \\
~ & Music & 2,000 & 60,382 \\
\midrule
\multirow{3}*{France (FR)}
~ & Books & 2,000 & 32,868 \\
~ & Dvd & 2,000 & 9,356 \\
~ & Music & 2,000 & 15,940 \\
\midrule
\multirow{3}*{Japanese (JA)}
~ & Books & 2,000 & 169,756 \\
~ & Dvd & 2,000 & 68,324 \\
~ & Music & 2,000 & 55,887 \\
\bottomrule
\end{tabular}
\caption{Statistics of examples in CLS datasets}\label{tab.CLS dataset}
\end{table}

\textbf{CoNLL}
We choose the CoNLL2002 \cite{tksintro2002conll} and CoNLL2003 \cite{tjongkimsang2003conll} datasets to test the performance of the model on the NER task. CoNLL2002 contains two languages, Spanish and Dutch, while CoNLL2003 contains English and Germany. English is used as the source language, and German, Spanish, and Dutch as the target languages during our experiments. As a dataset of NER task, there are four types of entity: ORG (Organization), LOC (Location), MISC (Misc), and PER (Person). This dataset provides the training set, validation set, and test set for each language. The training and validation set of English are used for training and evaluation, respectively. Test sets of other languages are used for testing. In the BFT process, we collect task-related unlabeled data directly from Wikipedia rather than using training sets in target languages as unlabeled data.
`
\begin{table}[htbp]
\small
\centering
\setlength{\tabcolsep}{4mm}{
\begin{tabular}{lcccc}
\toprule
\textbf{Language} & Train & Validate & Test\\
\midrule
English (EN) & 14,987 & 3,466 & 3,684 \\
German (DE) & 12,705 & 3,068 & 3,160\\
Spanish (ES) & 8,323 & 1,915 & 1,517\\
Dutch (NL) & 15,806 & 2,895 & 5,195\\
\bottomrule
\end{tabular}
}
\caption{Statistics of examples in CoNLL datasets}\label{tab.CoNLL dataset}
\end{table}

\begin{table*}[ht]
\centering
\small
\begin{tabular}{lccccccccccc}
\toprule
\multirow{2}*{\textbf{Model}} & \multirow{2}*{\textbf{w/o T/P}} 
~ & \multicolumn{3}{c}{\textbf{DE}} & \multicolumn{3}{c}{\textbf{FR}} &  \multicolumn{3}{c}{\textbf{JA}} & \multirow{2}*{\textbf{Average}}\\
\cmidrule(r){3-5}\cmidrule(r){6-8}\cmidrule(r){9-11}
 &  & \textbf{books} & \textbf{dvds} &  \textbf{music} & \textbf{books} & \textbf{dvds} &  \textbf{music} & \textbf{books} & \textbf{dvds} &  \textbf{music}  & \\ 
\midrule
Bi-PV & × & 79.51 & 78.60 & 82.45 & 84.25 & 79.60 & 80.09 & 71.75 & 75.40 & 75.45  & 78.57 \\
BiDRL & × & 84.14 & 84.05 & 84.67 & 84.39 & 83.60 & 82.52 & 73.15 & 76.78 & 78.77  & 81.34 \\
LASER & × & 84.15 & 78.00 & 79.15 & 83.90 & 83.40 & 80.75 & 74.99 & 74.55 & 76.30  & 79.47 \\
MonoX-PL & ${\surd}$ & 83.20 & 79.25 & 82.95 & 86.00 & 84.95 & 84.55 & 78.85 & 80.00 & 79.35  & 82.12 \\
MultiFit & ${\surd}$ & \textbf{89.60} & 81.80 & 84.40 & \textbf{87.84} & 83.50 & \textbf{85.60} & 80.45 & 77.65 & 81.50  & 83.59 \\
\midrule
MBERT & ${\surd}$ & 76.35 & 74.90 & 75.50 & 81.05 & 80.30 & 77.45 & 72.44 & 74.00 & 76.40  & 76.49 \\
Ours & ${\surd}$ & 87.65 & \textbf{85.70} & \textbf{86.40} & 86.80 & \textbf{86.40} & 83.10 & \textbf{83.29} & \textbf{83.00} & \textbf{84.55}  & \textbf{85.21} \\
\bottomrule
\end{tabular}
\caption{\label{tab.CLS_result} The accuracy of sentiment classification on CLS dataset. w/o T/P means this method doesn't use translation/parallel corpus}
\end{table*}

\begin{table*}[htbp]
\small
\centering
\setlength{\tabcolsep}{6.5mm}{
\begin{tabular}{lccccc}
\toprule
\textbf{Model} & \textbf{w/o T/P} & \textbf{DE} & \textbf{ES} & \textbf{NL} & \textbf{Average}\\ 
\midrule
\cite{wu2019beto} & ${\surd}$  & 69.56 & 74.96 & 78.60 & 74.37\\ 
\cite{WuQ2020Enhanced} & ${\surd}$  & 73.16 & 76.75 & 80.44 & 76.78\\
\cite{wu2020single} & × & 73.22 & 76.94 & 80.89 & 77.01\\
\cite{li2021cross} & × & 74.60 & 77.60 & 78.60 & 76.93\\
\cite{wu2020unitrans} & × & 74.82 & \textbf{79.31} & \textbf{82.90} & 79.01\\

\midrule
MBERT & ${\surd}$  & 72.24 & 74.96 & 79.31 & 75.50\\
Ours & ${\surd}$ & \textbf{76.88} & 78.38 & 82.08 & \textbf{79.11}\\
\bottomrule
\end{tabular}
}
\caption{\label{tab.CoNLL_result} The F1-score results on CoNLL dataset. }
\end{table*}
\subsection{Configuration}
\textbf{CLS} 
For the CLS dataset, the maximum length of input sentences is set to be 256.
In the Bilingual task fitting stage, We use an AdamW optimizer with a 0.1 warm-up rate and a 2e-5 maximum learning rate. The training batch size is 32. 
We use MBERT as our multi-lingual pre-trained model. Inspired by \cite{wu2019beto}, the parameters of its embedding layer and the bottom three transformer layers are frozen during training.
In the Self-training stage, the maximum learning rate is 1e-6. The training batch size is 30. The round number of iterations is 2, and the number of training epoch is 40 for each round. We use accuracy as the performance metric. For less time consumption, we randomly down-sample the unlabeled data to ~10,000 in each experiment. 

\textbf{CoNLL} 
For the CoNLL dataset, the maximum length of input sentences is set to be 128.
In the Bilingual task fitting stage, all the configurations are the same as those for the CLS dataset.
In the Self-training stage, we set the learning rate as 3e-5 for fine-tuning with soft labels, while 1e-6 with hard labels. The batch size is 32, and the number of training epoch is 10. The number of collected unlabeled data from Wikipedia is 60,000. 
In the training dataset, every word is tokenized and annotated with a BIO entity type. For each entity, the label of the first token is started with B-, while others are started with I-. Although each entity is tokenized into multiple tokens, we only use the prediction of the first token as a result for this entity. The performance metric is entity-level F1-score. 

\subsection{CLS Results}
We evaluate our method on the CLS dataset on the zero-shot learning task. In Table \ref{tab.CLS_result} we demonstrate the performance of our method and other methods, including MBERT, Bi-PV\cite{Bi-PV}, BiDRL\cite{BiDRL}, LASER\cite{Artetxe2018LASER}, MonoX-PL\cite{MonoX-PL} and MultiFit\cite{lample2019cross}. Among all of these methods, Bi-PV, BiDRL, and LASER use additional parallel corpus or pre-trained translation models, while MonoX-PL and MultiFit do not. Besides, in MultiFit, they use zero-shot predictions from a fine-tuned model based on LASER. From the table, we can see that our method brings great increases compared to MBERT (our baseline) and other methods and achieves the new SOTA. Although the results of MultiFit are better than ours on DE-books, FR-books, and FR-music, our method outperforms it on other configurations. Moreover, on average, we have a 1.62\% improvement compared to MultiFit.

\subsection{CoNLL Results}
Experimental results on the CoNLL dataset are shown in Table \ref{tab.CoNLL_result}. Our approach gets 76.88 in Germany, which outperforms other methods. In Spanish and Dutch, Wu's work \cite{wu2020unitrans} performs more superior, notwithstanding, they use an extra translation model. The F1-score of our method leaps by 3.095 over our baseline on these two languages. The average F1-score of our method pumps up by 3.61 compared to the MBERT.

%% file: content/Discussion.tex
\begin{table*}
\centering
\small
\begin{tabular}{lcccccccccc}
\toprule
\multirow{2}*{\textbf{Model}} 
~ & \multicolumn{3}{c}{\textbf{DE}} & \multicolumn{3}{c}{\textbf{FR}} &  \multicolumn{3}{c}{\textbf{JA}} & \multirow{2}*{\textbf{Average}}\\
\cmidrule(r){2-4}\cmidrule(r){5-7}\cmidrule(r){8-10}
  & \textbf{books} & \textbf{dvds} &  \textbf{music} & \textbf{books} & \textbf{dvds} &  \textbf{music} & \textbf{books} & \textbf{dvds} &  \textbf{music} & \\ 
\midrule
Ours & 87.65 & \textbf{85.7} & \textbf{86.40} & \textbf{86.80} & \textbf{86.40} & \textbf{83.10} & \textbf{83.29} & 83.00 & 84.55 & \textbf{85.21}\\
\midrule
- AT & \textbf{88.10} & 85.30 & 85.30 & 85.65 & 85.65 & 82.10 & 81.79 & \textbf{84.30} & \textbf{84.65} & 84.76 \\
- VM & 81.10 & 83.15 & 79.80 & 85.15 & 85.75 & 80.75 & 80.75 & 80.25 & 80.60 & 81.92 \\
- BTF & 81.65 & 80.20 & 78.55 & 82.25 & 82.20 & 79.00 & 81.45 & 77.40 & 78.75 & 80.16 \\
- ST & 87.65 & 83.00 & 84.25 & 85.55 & 85.15 & 82.65 & 82.69 & 82.45 & 83.05 & 84.05 \\
\midrule
MBERT & 76.35 & 74.90 & 75.50 & 81.05 & 80.30 & 77.45 & 72.44 & 74.00 & 76.4 & 76.49 \\
MBERT-wiki & 75.45 & 72.55 & 73.50 & 80.70 & 79.50 & 77.20 & 75.39 & 75.50 & 77.00 & 76.31 \\
\bottomrule
\end{tabular}
\caption{\label{tab.CLS_ablation} Ablation study evaluated on CLS dataset. AT, BFT and ST stand for Auto choose threshold, Bilingual Task Fitting and Self-training, respectively. MBERT-wiki is the model trained with BTF on the data randomly collected from Wikipedia.}
\end{table*}

\section{Discussion}

\begin{table}
    \small
    \centering
    \begin{tabular}{lcccc}
        \toprule
        \textbf{Model} & \textbf{DE} & \textbf{ES} & \textbf{NL} & \textbf{Average}\\ 
        \midrule
        Ours & \textbf{76.88} & \textbf{78.38} & \textbf{82.08} & \textbf{79.11}\\
        \midrule
        - AT & 76.23 & 78.08 & 81.54 & 78.62\\
        - VM & 76.76 & 77.51 & 81.39 & 78.55\\
        - BTF & 74.26 & 78.12 & 81.23 & 77.87\\
        - ST & 72.73 & 76.62 & 80.28 & 76.54\\
        \midrule
        MBERT & 72.24 & 74.96 & 79.31 & 75.50\\
        MBERT-wiki & 71.58 & 76.35 & 79.35 & 75.76\\
        \bottomrule
    \end{tabular}
    \caption{\label{tab.CoNLL_ablation} Ablation study evaluated on CoNLL dataset. AT, BFT and ST stand for Auto choose threshold, Bilingual Task Fitting and Self-training, respectively. MBERT-wiki is the model trained with BTF on the data randomly collected from Wikipedia.}
\end{table}

To investigate the effectiveness of each part in our method, we conducted an ablation study, and the problems in the two modules are analyzed experimentally.

\subsection{Ablation Study}

The experiments are divided into seven configurations as shown in Table \ref{tab.CLS_ablation} and Table \ref{tab.CoNLL_ablation}, with the corresponding modules removed from our method. We choose MBERT as the baseline. 
The tables demonstrate that: 1) Cross-lingual self-training module is critical to our method: the F1-score drops more than 2 points on the CoNLL dataset, and the accuracy drops 1 point on the CLS dataset when it is removed. 2) Bilingual Task Fitting may perform differently on different data or in different tasks. It improves the average accuracy by 7.55\% on the CLS dataset, mainly because the dataset provides a large amount of unlabeled data. The data is highly task-related and very similar to the training and test sets. In this way, the model can adapt to the task domain in the BTF process and obtain a better performance. 3) Auto choosing threshold and VM are helpful for cross-lingual transfer of the pre-trained model.

\subsection{Bilingual Task Fitting}

Most of the multi-lingual pre-trained language models are trained on universal corpus in a large set of languages. However, their performance on specific tasks and languages is limited. We try to improve the pre-trained model's effectiveness on specific tasks and languages by conducting MLM on the task-related data in the source and target languages. In Table \ref{tab.CLS_ablation}, we continue to pre-train the model on different datasets and compare the results. MBERT-wiki is the model trained with BTF on the data that are randomly collected from Wikipedia (same number as supported unlabeled dataset). We can see that the accuracy decreases by 0.18\%. However, by training on the task-oriented unlabeled data, as "Ours-ST" shows, BTF gets a great improvement of 7.56 points. For the CoNLL dataset, we also compared two different ways to collect unlabeled data. The first one is to select sentences from Wikipedia randomly, whereas the second one is to select sentences that include the named entities in the training data. In Table \ref{tab.CoNLL_ablation}, MBERT-wiki achieved a similar performance with MBERT (increase 0.26 F1-score), while "Ours-ST" outperforms MBERT (increase 1.04 F1-score). It shows that the dataset used for BTF is crucial, and when the task-relevant unlabeled data is not given, the data obtained by entity matching can still be considerably improved in the BTF phase.

\subsection{Cross-lingual Self-training}
In Self-training, the model is trained with the labels annotated by itself. Most prior cross-lingual self-training methods only do self-training for a single round. Unlikely, we consider our method as an iterative training process and carry out some experiments to verify whether Multi-rounds self-training helps improve the results persistently.

First, we use soft labels to perform multiple-rounds self-training on the CoNLL dataset. It is found that the results are almost converging in the first round with sufficient training (the configuration is shown in 4.2). Multiple-rounds self-training does not further improve the performance markedly, as shown in Table \ref{tab.soft_label_multi_round}.

\begin{table}[!ht]
    \small
    \centering
    \begin{tabular}{ccccc}
        \toprule
        \textbf{LAN} & \textbf{MBERT} & \textbf{Round1} & \textbf{Round2} & \textbf{Round3} \\ 
        \midrule
        DE & 72.73 & 75.22 & \textbf{75.43} & 75.31\\
        ES & 76.62 & \textbf{77.21} & 77.05 & 77.08\\
        NL & 80.28 & 81.10 & 80.97 & \textbf{81.17}\\
        \bottomrule
    \end{tabular}
    \caption{\label{tab.soft_label_multi_round} Multiple-rounds self-training with only soft labels on CoNLL dataset}
\end{table}

Then, we replace soft labels with hard labels. We find that multiple-rounds self-training can improve performance in general. However, it is unstable before reaching convergence. It is mainly because the pseudo hard labels usually contain a large amount of noise. In \cite{2021Self}, they find that filtering hard labels by a confidence threshold can effectively improve the results. Moreover, the selections of this threshold are always manual. Many groups of experiments have to be made to choose appropriate thresholds, making the experimental process much more complex.

Therefore, to solve the problems above, we use both soft labels and hard labels during self-training, as mentioned in 3.2. Table \ref{tab.soft_label_multi_round} shows one-round training with soft labels in our method improves the performance stably. Furthermore, we enhance its effectiveness by autonomously selecting a threshold to filter the hard labels.


\begin{figure}[!ht]
    \tiny
    \centering
	\includegraphics[width=0.47\textwidth]
	{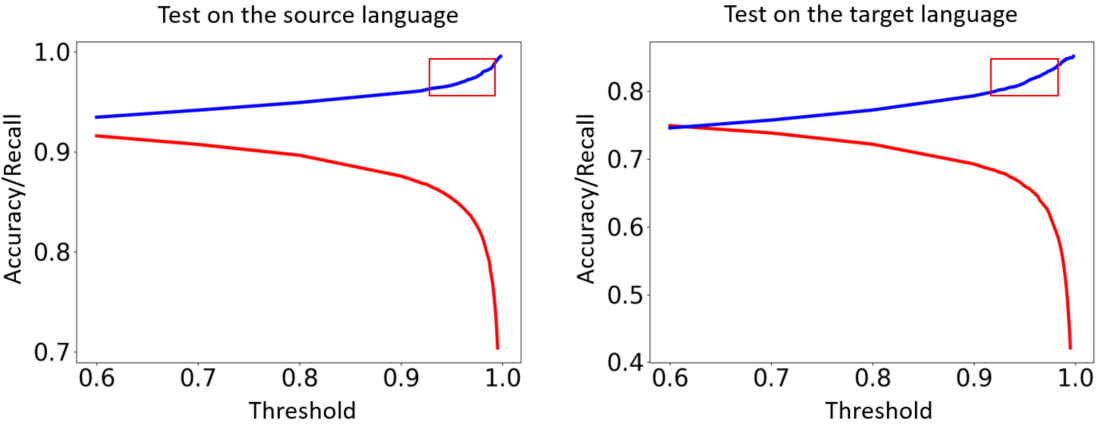}
	\caption{The relationship between accuracy and recall when threshold changes on the data in source and target languages. The blue curve represents for accuracy while the red curve stands for recall.}
	\label{Fig.th}
\end{figure}

\begin{table}[!ht]
    \tiny
    \centering
    \begin{tabular}{ccccccc}
        \toprule
        \textbf{LAN} & \textbf{Origin} & \textbf{0.0} & \textbf{0.9} & \textbf{0.99} & \textbf{0.999} & \textbf{Ours} \\ 
        \midrule
        DE & 75.22 & 76.15 & 76.38 & 76.24 & 75.87 & 76.40\\
        ES & 77.21 & 77.83 & 77.51 & 77.91 & 76.96 & 77.79\\
        NL & 81.10 & 81.05 & 82.08 & 81.77 & 81.44 & 81.62\\
        \bottomrule
    \end{tabular}
    \caption{\label{tab.th_compare} Performance with different thresholds on CoNLL dataset}
\end{table}

In cross-lingual self-training, it is essential to ensure that the hard labels have high accuracy while the data has high diversity. To strike a balance between them, we experiment and observe the relationship between accuracy and recall with different thresholds. Figure \ref{Fig.th} shows that the accuracy increases too while the recall drops with the increasing of the threshold. Besides, it also shows that the relationships are pretty similar in both the source language and target language. Since in zero-shot cross-lingual transfer, we only have labels for training data in the source language, the threshold is selected on the source language and directly used on the target language. As described in 3.2, we select the threshold by looking at the point where the accuracy curve rate is the most enormous. We compare the results of using fixed thresholds and our obtained thresholds in one round on CoNLL data, and they are shown in Table \ref{tab.th_compare}. We can see that too low or too high thresholds will lead to poor performance. However, using the threshold obtained by our method, the model performs better than most other manually set thresholds.

Another thing about using pseudo hard labels for cross-lingual self-training is that some predicted hard labels of training data in the source language are inconsistent with their corresponding labels. These samples cannot be correctly predicted by the self-trained model, which indicates that the knowledge learned by the model through these samples is not retained during the knowledge transfer process. Moreover, these samples may disturb the self-training and lead to worse performance. Therefore, in every round of self-training, we permanently remove these samples and retain the training corpus in the source language only when their predictions are consistent with the original labels. Then the consistent data and the data with the pseudo hard labels in the target language  are combined for subsequent fine-tuning.

\begin{table}[!ht]
    \tiny
    \centering
    \begin{tabular}{ccccccc}
        \toprule
        \textbf{LAN} & \textbf{MBERT} & \textbf{Soft1} & \textbf{Hard1} & \textbf{Soft2} & \textbf{Hard2} & \textbf{Soft3} \\ 
        \midrule
        DE & 72.73 & 75.22 & 76.40 & 76.88 & 76.68 & 76.88\\
        ES & 76.62 & 77.21 & 77.79 & 78.38 & 78.23 & 78.55\\
        NL & 80.28 & 81.10 & 81.62 & 82.08 & 81.52 & 81.93\\
        \bottomrule
    \end{tabular}
    \caption{\label{tab.soft_hard_label_multi_round} Multi-round self-training with soft labels and hard labels on CoNLL dataset}
\end{table}

Another important hyperparameter for self-training is the iteration round N. As shown in Table \ref{tab.soft_hard_label_multi_round}, we find that the model is close to being converged on CoNLL after three iterations. Therefore, for the CoNLL dataset, the N is set to be 1.5 (Soft-Hard-Soft). As for the CLS dataset, the N is 2 according to our experimental results.


%% file: content/Conclusion.tex
\section{Conclusion}
We propose a novel framework to conduct zero-shot cross-lingual tasks without using any parallel corpus or translation models. It is a two-stage approach constructed by a Bilingual Task Fitting and Self-training module. The Bilingual Task Fitting module enhances the performance of pre-trained multi-lingual language models on tasks in both the source and target languages. Then, the proposed self-training module improve the results further.
Our proposed method performs well on different tasks, including multi-lingual NER and Sentiment classification. It achieves new SOTA on both tasks. 
In future work, we will try to combine two stages of our method more effectively and transfer knowledge from multiple source languages. 